\pdfoutput=1
\documentclass[11pt]{article}
\usepackage{amta2010}
\usepackage{times}
\makeatletter
\newcommand{\@BIBLABEL}{\@emptybiblabel}
\newcommand{\@emptybiblabel}[1]{}
\makeatother
\usepackage{url}
\usepackage{latexsym}
\usepackage[pdftex]{graphicx}
\usepackage[small,compact]{titlesec}
\usepackage{enumitem}
\usepackage{hyperref}
\title{Semantically-Informed Syntactic Machine Translation:\\ A Tree-Grafting
Approach}

\author{\mbox{~}
Kathryn Baker \\
U.S. Dept. of Defense \\
Fort Meade, MD 20755 \\
{\small\tt  klbake4@tycho.ncsc.mil}
\And Michael Bloodgood \\
JHU HLTCOE\\
Baltimore, MD 21218 \\
{\small\tt bloodgood@jhu.edu} 
\And Chris Callison-Burch \\
JHU HLTCOE\\
Baltimore, MD 21218 \\
{\small\tt ccb@cs.jhu.edu} 
\And Bonnie J. Dorr \\
University of Maryland \\
College Park, MD 20742\\ 
{\small\tt bonnie@umiacs.umd.edu}
\AND Nathaniel W. Filardo \\
JHU HLTCOE \\
Baltimore, MD 21211\\ 
{\small\tt nwf@cs.jhu.edu}
\And Lori Levin \\
CMU \\
Pittsburgh, PA 15213\\ 
{\small\tt lsl@cs.cmu.edu} 
\And Scott Miller \\
 BBN \\
Cambridge, MA 02138\\ 
{\small\tt smiller@bbn.com\mbox{~~}}
\And Christine Piatko \\
JHU/APL \\
Laurel, MD 20723\\
{\footnotesize\tt christine.piatko@jhuapl.edu}
}

\date{}

\begin{document}
\maketitle
\begin{abstract}

We describe a unified and coherent syntactic framework for supporting
a semantically-informed syntactic approach to statistical machine translation. 
Semantically enriched syntactic tags assigned to the target-language
training texts improved translation quality. 
The resulting system significantly outperformed a
linguistically naive baseline model (Hiero), and reached the highest scores
yet reported on the NIST 2009 Urdu-English translation task.  
This finding supports the hypothesis (posed by many researchers in
the MT community, e.g., in DARPA GALE) that both
syntactic and semantic information are critical for improving
translation quality---and further demonstrates that large gains
can be achieved for low-resource languages 
with different word order than English.
\end{abstract}

\section{Introduction}

This paper describes a tree-grafting approach to incorporating
named entities and modality into a 
unified and coherent syntactic framework, as a first
step toward supporting Semantically-Informed Machine Translation (SIMT). 
The implementation of this approach was the result of a large effort
undertaken in the summer of 2009.
The most significant result of the SIMT 
effort
was the integration of
semantic knowledge into statistical machine translation in a unified
and coherent syntactic framework.  By augmenting hierarchical phrase-based
translation rules with syntactic labels that were 
extracted from a parsed parallel corpus, and further augmenting the
parse trees with semantic elements such as named-entity markers and
modality (through a process we refer to as {\it grafting\/}), 
we produced a better model for translating Urdu and
English. The resulting system significantly outperformed the
linguistically naive baseline Hiero model, and reached the highest scores
yet reported on the NIST 2009 Urdu-English translation task.

We note that, while our largest gains were from syntactic enrichments
to the model, smaller (but significant) gains were achieved by
injecting semantic knowledge into the syntactic paradigm.  Of course,
entities and modalities are only a small piece of the much larger
semantic space, but demonstrating success on these new, unexplored
semantic aspects of language bodes well for (larger) improvements 
based on the incorporation of
other semantic aspects (e.g., relations and temporal knowledge).
Moreover, we believe this syntactic framework to be well suited for further exploration of the impact of many different types of semantics on MT quality. 
Indeed, it would not have been possible to
initiate the current study without the foundational work that gave
rise to a syntactic paradigm that could support these semantic
enrichments.  We believe this framework will be especially useful for
exploring other languages with few resources and different word order than English.

\begin{figure*}[t]
\begin{center}
\includegraphics[width=\linewidth]{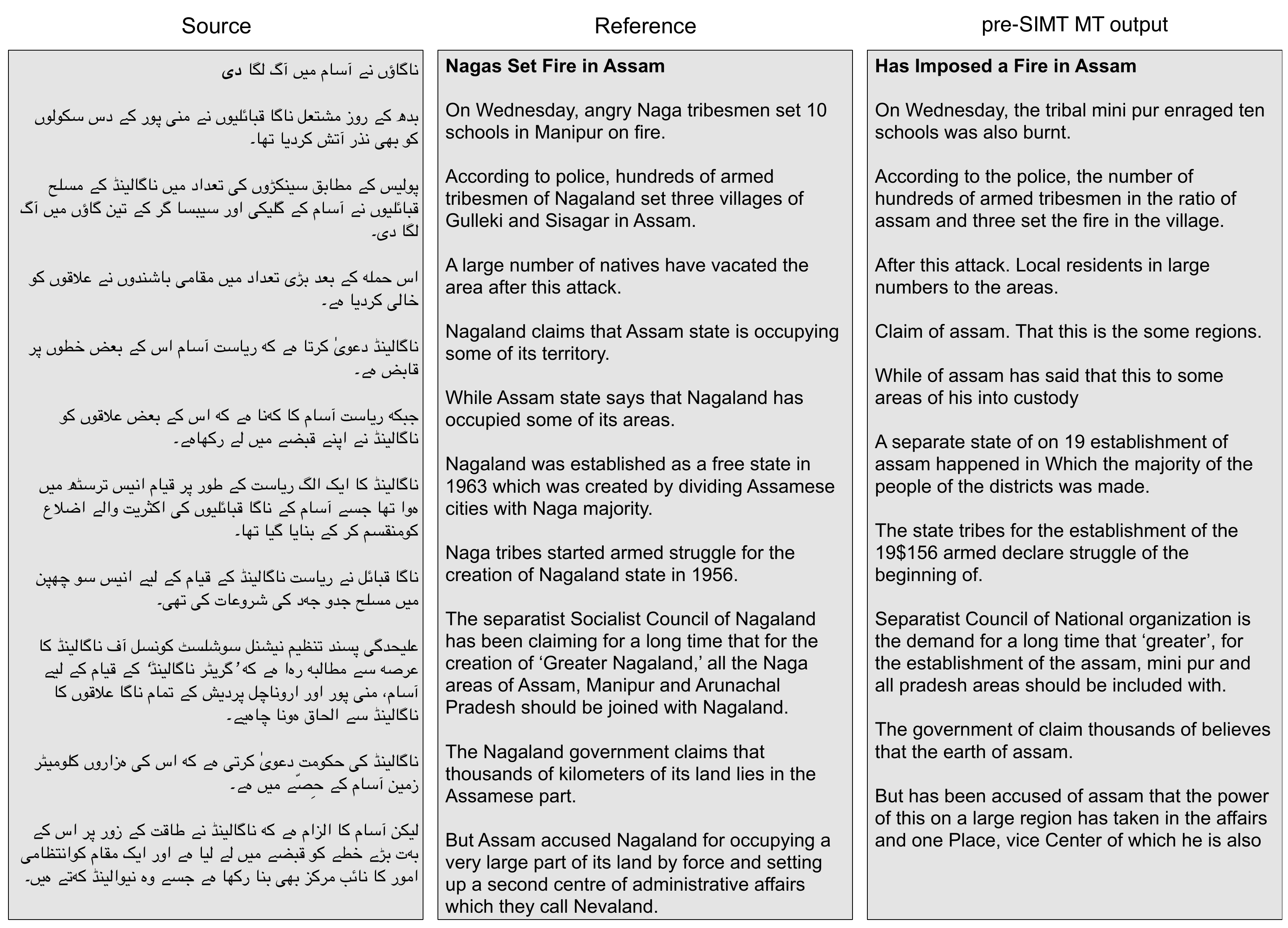}
\end{center}
\vspace*{-0.5cm}
\caption{An example of Urdu-English translation.  Shown are an Urdu
source document, a reference translation produced by a professional
human translator, and machine translation output from a phrase-based model (Moses) without linguistic information, which is representative of state-of-the-art 
MT quality before the SIMT effort.}
\label{Urdu-example}
\end{figure*}

The semantic units that we examined 
in this effort 
were named entities (such as people or organizations) and
modalities (indications that a statement represents something that has
taken place or is a belief or an intention).  Other 
semantic units
such as
relations between entities and events, were not part of 
this effort,
but we believe they could be similarly incorporated into the
framework.  We chose to examine 
semantic units
that canonically exhibit two
different syntactic types: nominal, in the case of named entities, and
verbal, in the case of modality.

Named entities have been the focus of information extraction research
since the Message Understanding Conferences of the
1980s~\cite{DBLP:conf/coling/GrishmanS96}.  Automatic taggers identify
semantic types such as person, organization, location, date, facility,
etc.  In this research effort we tagged English documents using an
HMM-based tagger derived from Identifinder~\cite{Bikel:1999}.

Modality is an extra-propositional component of meaning.  In {\it John
may go to NY\/}, the basic proposition is {\it John go to NY\/} and
the word {\it may\/} indicates modality. Van der Auwera and
Amman~\shortcite{VanDerAuweraAmman} define core cases of modality:
{\it John must go to NY\/} (epistemic necessity), {\it John might go to NY\/}
(epistemic possibility), {\it John has to leave now\/} (deontic necessity) and
{\it John may leave now\/} (deontic possibility). Many
semanticists~\cite{Kratzer,VonFintelIatridou} define modality as
quantification over possible worlds. {\it John might go\/} means that there
exist some possible worlds in which John goes. Another view of
modality relates more to a speaker's attitude toward a
proposition~\cite{NirenburgMcShane,McShaneEtAl:2004}.
Modality resources built for this purpose have been 
described previously~\cite{LRECModality:2010}.  

\begin{figure*}[t]
\begin{center}
\includegraphics[height=3in]{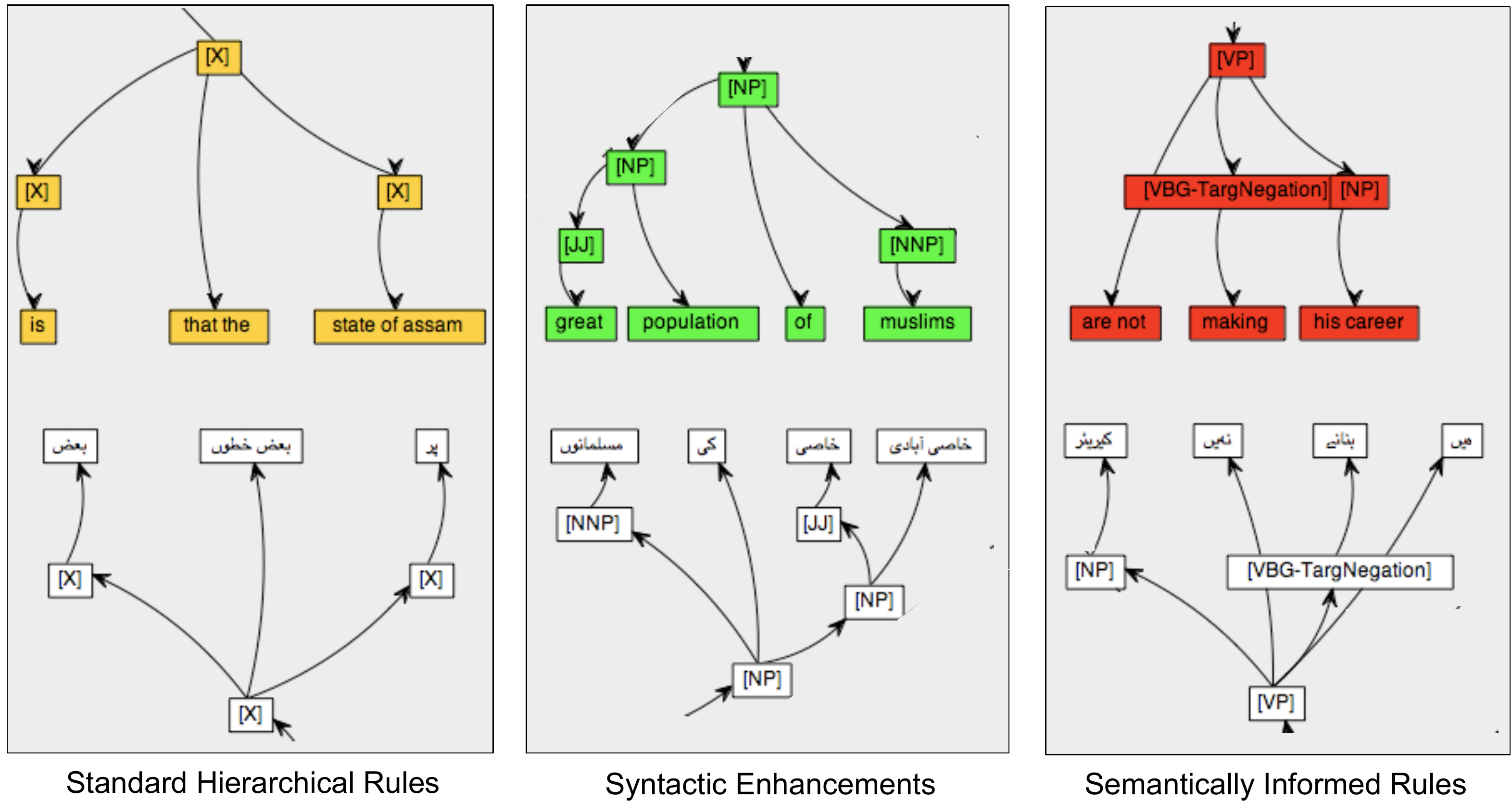}
\end{center}
\vspace*{-0.5cm}
\caption{The evolution of a semantically informed approach to our synchronous context free grammars (SCFGs). 
At the start of summer the decoder used translation rules with a single generic non-terminal symbol, later syntactic categories were used, and 
by the end of the summer the translation rules included semantic elements such as named entities and modalities.}
\label{Evolution-to-SIMT-Joshua}
\end{figure*}

This paper will focus on a tree-grafting mechanism used to enrich the
machine-translation output and on the resulting improvements to
translation quality when the training process for the
machine-translation systems included tagging of named entities and
modality.

The next section provides the motivation behind the SIMT 
approach.  Section~\ref{joshua}
presents implementation details of the semantically-informed
syntactic system.  Section~\ref{algorithm} describes the tree grafting
algorithm.  Section~\ref{results} provides the results of this
work. Following this, Section~\ref{related-work} examines work that is
related to our approach.  Finally, Section~\ref{conclusions} presents
conclusions and future work.

\section{Motivation}
\label{motivation}

The aim of the SIMT 
effort
was to
provide a generalized framework for representing structured semantic
information, such as named entities and modality, and to investigate
whether incorporating this sort of information into machine
translation (MT) systems could produce better translations.  The SIMT
effort
differs from other efforts in MT, most notably the DARPA Global
Autonomous Language Exploitation (GALE) initiative, in at 
least two ways:
\begin{enumerate}[topsep=0pt, partopsep=0pt, itemsep=0.5pt, parsep=0.5pt]
\item The SIMT 
effort
worked on translation for a low-density language,
with a minimal amount of bilingual training data.  In GALE,
hundreds of millions of words worth of bilingual texts are used to
train statistical translation models.  
In the SIMT effort,
only 1.7 million
words of Urdu-English texts were available. 
Table~\ref{data-set-sizes} provides the data set sizes used in our experiments.
\item The SIMT effort
showed significant improvements from incorporating syntax and semantics
into machine translation, whereas syntactic
translation models have not shown dramatic improvements in GALE's
Arabic-English translation task.  The
improvements for Urdu translation described here are probably due to the fact that it is a low-resource, verb-final language and 
so requires generalization beyond phrase-based or hierarchical phrase-based models.
\end{enumerate}
These differences created novel research directions for 
our
effort,
and resulted in promising findings that suggest that both
syntactic and semantic information are critical for improving
translation quality.

\begin{table}[b]
\begin{tabular}{lrrrrr}
\hline
\multicolumn{2}{c}{} & \multicolumn{2}{c}{Urdu} & \multicolumn{2}{c}{English} \\
set & lines &  tokens &  types &  tokens & types\\
\hline 
training & 202k & 1.7M & 56k & 1.7M & 51k \\
dev & 981 & 21k & 4k & 19k & 4k \\
devtest & 883 & 22k & 4k & 19-20k & 4k \\
test & 1792 & 42k & 6k & 38-41k & 5k \\
\end{tabular}
\vspace*{-0.5cm}
\caption{The size of the various data sets used for the experiments in this paper including the training, development (dev), 
incremental test set (devtest) and blind test set (test).  The dev/devtest was a split of the NIST08 Urdu-English test set, 
and the blind test set was NIST09.}
\label{data-set-sizes}
\end{table}

It is informative to look at an example translation to understand the
challenges of translating important semantic entities when working
with a low-resource language pair. Figure~\ref{Urdu-example} shows an
example taken from the 2008 NIST Urdu-English translation task, and
illustrates the translation quality of a state-of-the-art Urdu-English
system 
(prior to the SIMT effort).
The small amount of training data for this language pair (see Table~\ref{data-set-sizes}) results
in significantly degraded translation quality compared, e.g., to an
Arabic-English system that has more than 100 times the amount of
training data.  

The machine translation output in
Figure~\ref{Urdu-example} was produced using Moses~\cite{Moses}, a state-of-the-art
phrase-based machine translation system that by default does not incorporate any
linguistic information (e.g., syntax or morphology or transliteration
knowledge).  As a result, words that were not
directly observed in the bilingual training data were untranslatable.  
Names, in particular, are problematic.
 For example, the lack of translation for {\it Nagaland} and {\it Nagas}  induces
 multiple omissions throughout the translated text.   This is because out of vocabulary words are deleted from the Moses output. 

\begin{figure}
\begin{center}
\includegraphics[height=1.25in]{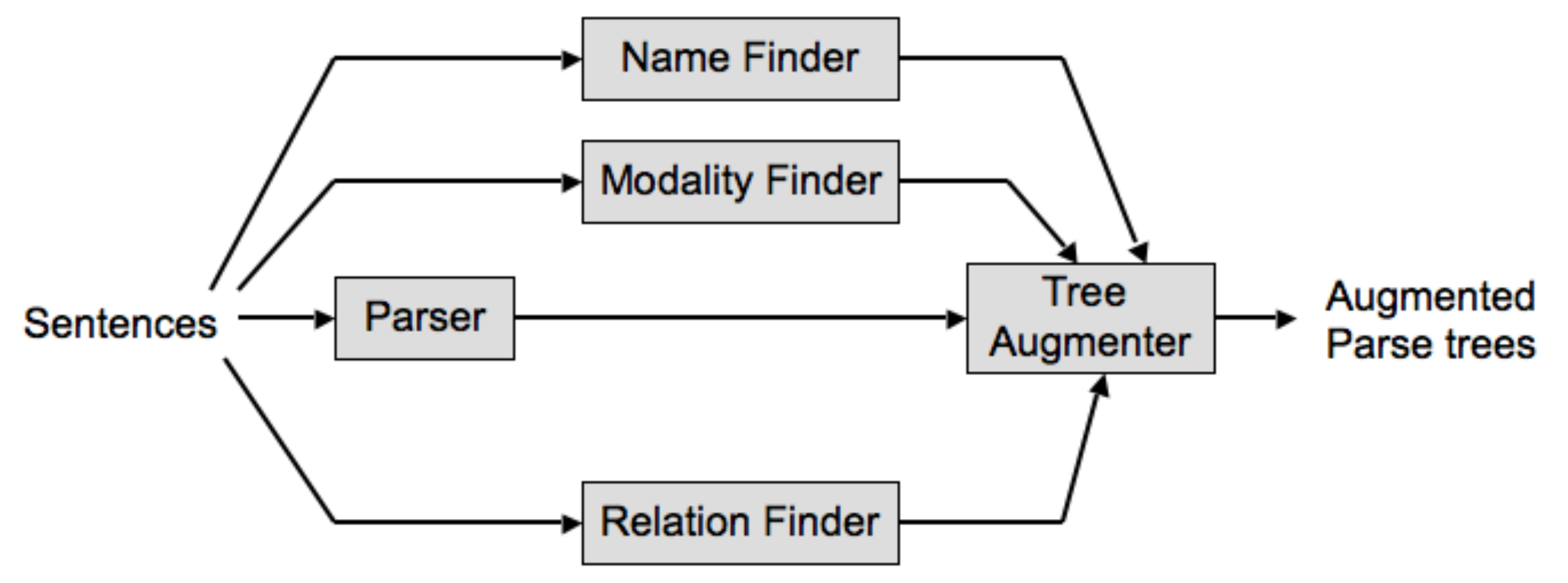}
\end{center}
\vspace*{-0.5cm}
\caption{Workflow for producing semantically-grafted parse trees.  The English side of the parallel corpus is automatically parsed, 
and also tagged with modality and named-entity markers.  These tags are then grafted onto the syntactic parse trees.  
The relation finder was designed for additional tagging but was not implemented in the current work.  
(Future work will test relations as another component of meaning that may contribute toward improved MT ouput.)}
\label{method-for-producing-hive-augmented-parse-tree}
\end{figure}

We use modality and named-entity
tags as higher-order symbols inside the translation rules used
by the translation models.  Generic symbols in translation rules (e.g.,
the
non-terminal symbol ``X'') were replaced with structured
information at multiple levels of abstraction, using a tree-grafting
approach, as described in more detail in the following sections.
Figure~\ref{Evolution-to-SIMT-Joshua}
illustrates the evolution of the translation rules that we used, first replacing ``X'' with grammatical categories and then with
semantic categories.

\section{Tree-Grafting to refine translation grammars with semantic categories}
\label{joshua}

We use synchronous context free grammars (SCFGs) as the underlying
formalism for our statistical models of translation.  SCFGs provide a
convenient and theoretically grounded way of incorporating linguistic
information into statistical models of translation, by specifying
grammar rules with syntactic non-terminals in the source and target
languages.  We refine the set of non-terminal symbols so that they not
only include syntactic categories, but also semantic categories.

\newcite{Chiang:2005} re-popularized the use of SCFGs for machine
translation, with the introduction of his hierarchical phrase-based
machine translation system, Hiero.  Hiero uses grammars with a single
non-terminal symbol ``X'' rather than using linguistically informed
non-terminal symbols.  When moving to linguistic grammars, we use the
Syntax Augmented Machine Translation (SAMT) developed by
\newcite{Venugopal:2007}.  In SAMT the ``X'' symbols in translation
grammars are replaced with nonterminal categories derived from parse
trees that label the English side of the Urdu-English parallel
corpus.\footnote{For non-constituent phrases, composite CCG-style
  categories are used~\cite{Steedman:1999}.}  We refine the syntactic
categories by combining them with semantic categories.  This
progression is illustrated in Figure~\ref{Evolution-to-SIMT-Joshua}.

We extracted SCFG grammar rules containing named entities and modality
using an extraction procedure that requires parse trees for one side
of the parallel corpus. While it is assumed that these trees are
labeled and bracketed in a syntactically motivated fashion, the
framework places no specific requirement on the label inventory.  We
take advantage of this characteristic by providing the rule extraction
algorithm with augmented parse trees containing syntactic labels that
have named entities and modalities grafted onto them so that they
additionally express semantic information.

\begin{figure*}
\begin{center}
\includegraphics[width=\linewidth]{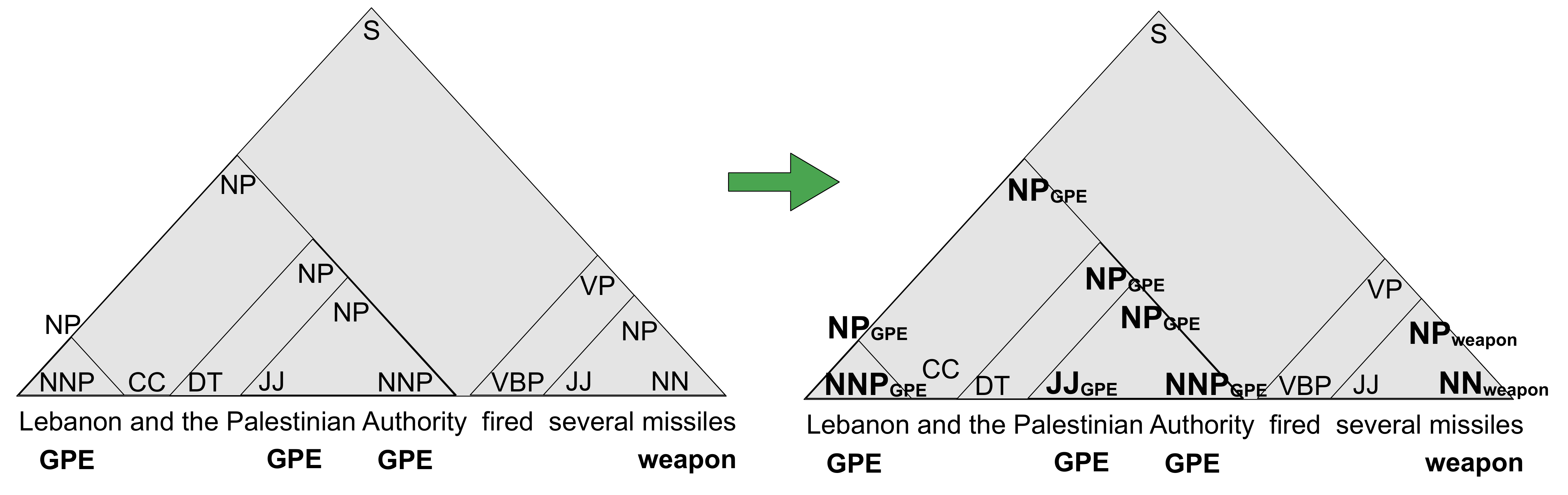}
\end{center}
\vspace*{-0.5cm}
\caption{A sentence on the English side of the bilingual parallel training corpus is parsed with a syntactic parser, and also tagged with a named entity tagger. 
The tags are then grafted onto the syntactic parse tree to form new categories like NP-GPE and NP-weapon.  
Grafting happens prior to extracting translation rules, which happens normally except for the use of the augmented trees.}
\label{Parse-Tree-for-Sample-Sentence}
\end{figure*}

Our strategy for producing semantically-grafted 
parse trees involves three steps:  
\begin{enumerate}[topsep=0pt, partopsep=0pt, itemsep=0.5pt, parsep=0.5pt]
\item The English sentences in the parallel training data are parsed with a syntactic parser.  In our work, we used the lexicalized probabilistic context free 
grammar parser provided by Basis Technology Corporation.
\item The English sentences are named-entity-tagged by the Phoenix tagger~\cite{richman-schone:2008:ACLMain} and modality-tagged by the 
system described in~\cite{LRECModality:2010}.
\item The named entities and modalities are grafted onto the syntactic parse trees using a tree-grafting procedure. 
The grafting procedure was implemented as a part of the SIMT effort.  Details are spelled out further in Section~\ref{algorithm}.
\end{enumerate}
The workflow for producing semantically-grafted trees is illustrated in Figure~\ref{method-for-producing-hive-augmented-parse-tree}.
Figure~\ref{Parse-Tree-for-Sample-Sentence} illustrates how named-entity tags are grafted onto a parse tree.  
We note that while our framework is general, 
we focus the discussion here on the particular semantic elements (named entities and modalities) that were incorporated during the 
SIMT effort.

Once the semantically-grafted
trees have been produced for the parallel corpus, the
trees are presented, along with word alignments (produced by an
aligner such as GIZA++), to the rule extraction software to extract
synchronous grammar rules that are both syntactically and semantically
informed.  These grammar rules are used by the decoder to produce
translations.  In our experiments, we used the Joshua decoder
\cite{Li:2009}, the SAMT grammar extraction software
\cite{Venugopal2009}, and special purpose-built tree-grafting software.

\begin{figure}[t]
\begin{center}
\includegraphics[width=\linewidth]{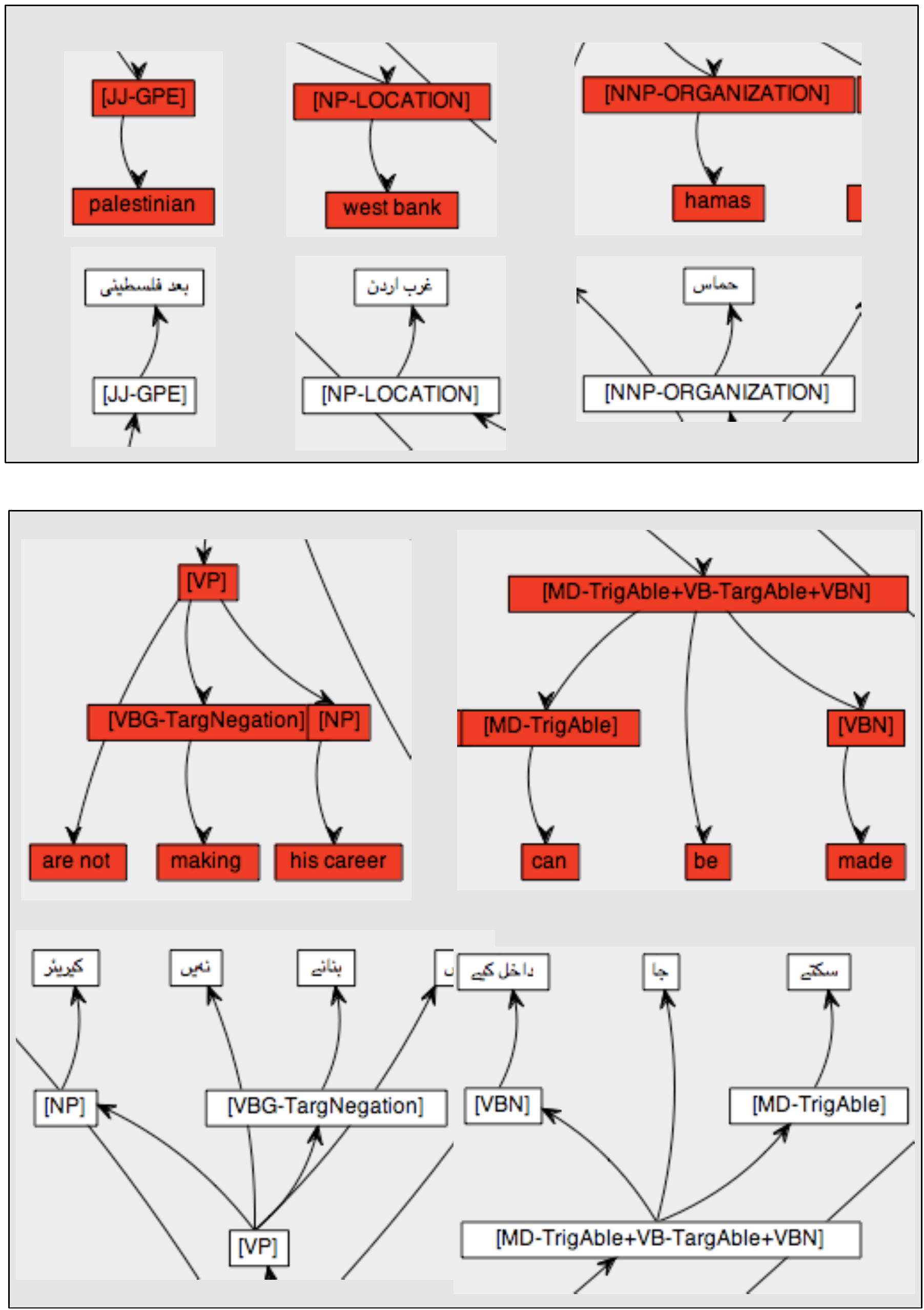}
\vspace*{-0.5cm}
\caption{Example translation rules with named entity tags and modalities combined with syntactic categories.}\label{derivation-with-modalities}
\end{center}
\end{figure}

Figure~\ref{derivation-with-modalities} shows example semantic rules
that are used by the decoder.  The noun-phrase rules are augmented
with named entities, and the verb phrase rules are augmented with
modalities.  The semantic categories are listed in
Table~\ref{NEtable} and Table~\ref{Modalitytable}.  Because these get
marked on the Urdu source as well as the English translation,
semantically enriched grammars also act as very simple named entity or
modality taggers for Urdu.  However, only entities and modalities that
occurred in the parallel training corpus are marked in the output.

\section{Tree-Grafting Algorithm}
\label{algorithm}

The overall scheme of our tree-grafting algorithm is to
match semantic tags to syntactic categories. 
There are two inputs to the process.  Each is derived from a common
text file of sentences.  The first input is a list of standoff
annotations for the 
semantic units
in the input sentences, indexed by sentence
number.  The second is a list of parse trees for the sentences in Penn
Treebank format,
indexed by sentence number.

Table~\ref{NEtable} lists the entity types identified during
the SIMT effort, with examples.  Table~\ref{Modalitytable} likewise lists the modality types that were produced by the modality tagger. 
\newcite{LRECModality:2010} described a system that automatically tags triggers and targets of modality.  
A trigger is a word with a modal meaning like {\it believe\/}, {\it possible\/}, or {\it want\/}.  
A target is a word in the scope of the trigger.
For example, the sentence  {\it The students are able to swim\/} is tagged 
as {\it The students are  $\langle$TRIG-ABLE able$\rangle$ to $\langle$TARG-ABLE to swim$\rangle$\/}.

\begin{table}[t]
\begin{center}
\begin{small}
\begin{tabular}{l l}\\
{\bf Named Entity} & {\bf Example} \\
\hline
AGE & 50 years old \\
DATE & September 26, 2009\\
FACILITY & Southwestern Medical Center \\
GPE (Geo-political entity) & New York\\
GPE-ite & Australian\\
LOCATION & West Sea \\
MONEY & 15,000 pounds \\ 
OCCUPATION & governor \\
ORGANIZATION & United Nations \\
ORGANIZATION-ite & marines \\
PERCENT & 3.1 percent \\
PERSON & Tony Blair \\
TIME & 2030 GMT \\
\end{tabular}
\end{small}
\end{center}
\vspace*{-0.5cm}
\caption{Named entity tags}
\label{NEtable}
\begin{center}
\begin{small}
\begin{tabular}{ll}\\
Require& NOTPermit\\
Permit& NOTRequire\\
Succeed & NOTSucceed\\
SucceedNegation & NOTSucceedNegation\\
Effort& NOTEffort\\
EffortNegation & NOTEffortNegation\\
Intend& NOTIntend\\
IntendNegation& NOTIntendNegation\\
Able& NOTAble\\
AbleNegation& NOTAbleNegation\\
Want& NOTWant\\
Belief& NOTBelief\\
Firm\_Belief& NOTFirm\_Belief\\
Negation & \\
\end{tabular}
\end{small}
\end{center}
\vspace*{-0.5cm}
\caption{Modality tags with their negated versions}
\label{Modalitytable}
\end{table}

The tree-grafting algorithm proceeds as follows. For each sentence, we
iterate over the list of semantic tags.  For each semantic tag, we
determine the parent node or nodes in the corresponding syntactic
parse tree that dominate the word sequence covered by the tag.  The
following tests are then applied:
\begin{itemize}[topsep=0pt, partopsep=0pt, itemsep=0.5pt, parsep=0.5pt]
\item If the semantic and syntactic units correspond exactly, graft
  the name of the semantic tag onto the highest corresponding
  syntactic constituent in the tree.  For example, in
  Figure~\ref{Parse-Tree-for-Sample-Sentence}, the NNP ``Lebanon''
  receives a GPE (geo-political entity) tag at the NP constituent level.
\item For the case of named entities: If the semantic tag corresponds
  to words that are adjacent daughters in a syntactic constituent,
  but less than the full constituent, insert an NP node dominating
  those words into the parse tree, as a daughter of the original
  syntactic constituent.  The name of the semantic tag is grafted onto
  the new NP node.  This is a case of rule splitting.
\item If a syntactic constituent selected for grafting has already
  been labeled with a semantic tag, overlay that tag.
\item If the words covered by the semantic tag fall across two
  syntactic constituents, do nothing.  This is a case of crossing
  brackets.
\end{itemize}

\begin{figure}[t]
\begin{center}
\includegraphics[width=\linewidth]{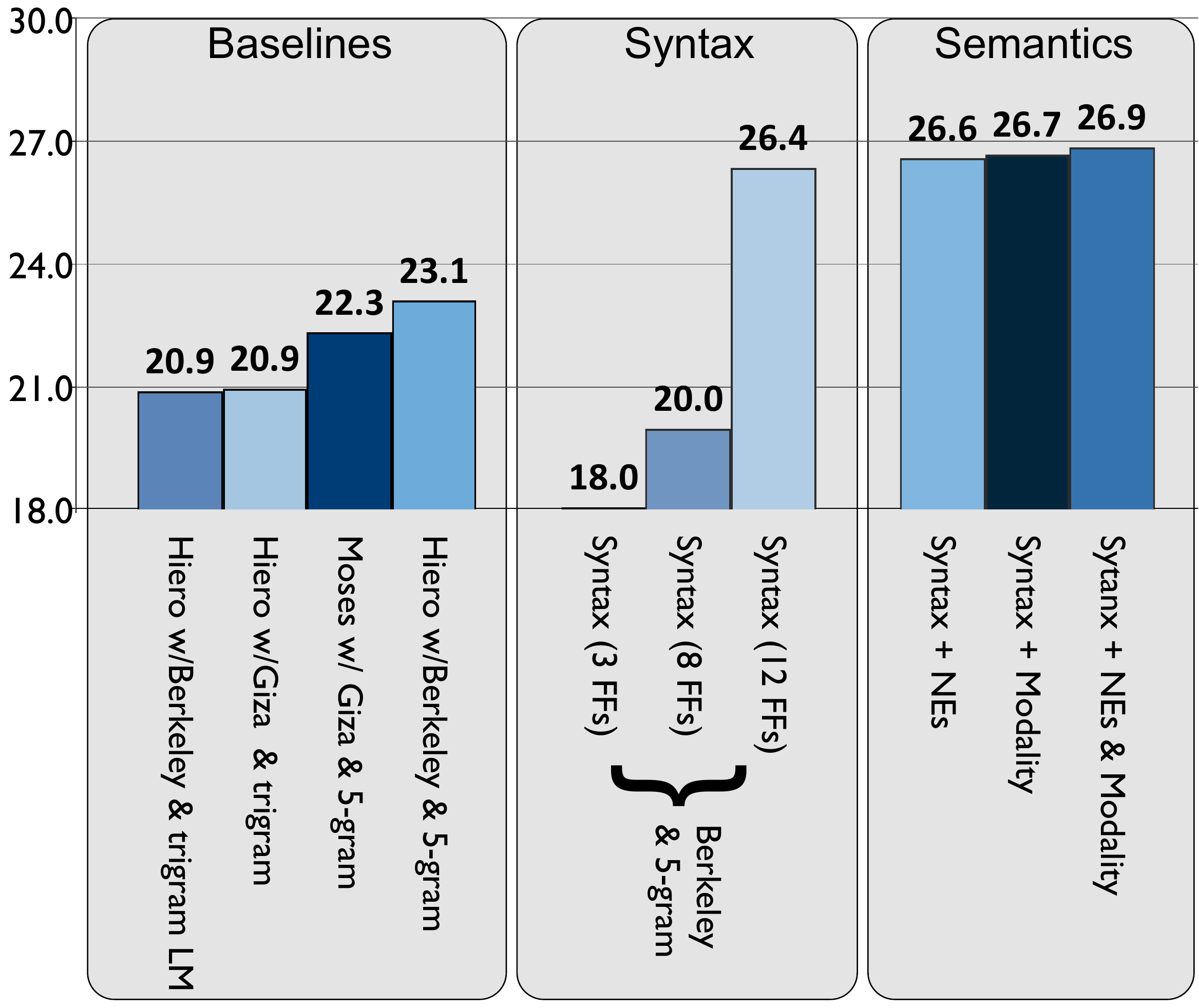}
\vspace*{-0.5cm}
\caption{Results for a range of experiments conducted during the 
SIMT effort. Results show scores for baseline systems, which here include a phrase-based model (Moses) and a hierarchical phrase-based model (Hiero), neither of 
which make use of syntactic information. 
These also show the substantial improvements when syntax is introduced, along with different numbers of feature functions (FFs), 
and further improvements from semantic elements. The scores are lowercased Bleu calculated on the held-out devtest set.}
\label{joshua-experiments}
\end{center}
\end{figure}

Our tree-grafting procedure was simplified to accept a single semantic
tag per syntactic tree node as the final result.  The algorithm keeps
the last tag seen as the tag of precedence.  In practice, we
established a precedence ordering for modality tags over named entity
tags by grafting named entity tags first and modalities second.  Our
intuition was that, in case of a tie, finer-grained verbal categories
would be more helpful to parsing than finer-grained nominal
categories.\footnote{In testing we found that grafting named entities
  first and modalities last yielded a slightly higher Bleu score than
  the reverse order.}  In case a word was tagged both as a modality
target and a modality trigger, we gave precedence to the target tag.
This is because, while modality targets vary, modality triggers are
generally identifiable with lexical items.  Finally, we used a
simplified specificity ordering of modality tags, borrowing from an
approach described in~\cite{LRECModality:2010}, to ensure precedence
of more specific tags over more general
ones. Table~\ref{Modalitytable} lists the modality types from highest
(Require modality) to lowest (Negation modality)
precedence.\footnote{Future work could include exploring additional
  methods of resolving tag conflicts or combining tag types on single
  nodes, e.g. by inserting multiple intermediate nodes (effectively
  using unary rewrite rules) or by stringing tag names together.}

\section{Results}
\label{results}

Figure~\ref{joshua-experiments} gives the results for a number of
experiments conducted during the 
SIMT effort.\footnote{These experiments
  were conducted on the devtest set, containing 883 Urdu sentences (21,623 Urdu words) and four reference translations per sentence.  The Bleu score for these
  experiments is measured on uncased output, which in general should
  be higher, but the devtest effectively had only three reference
  translations. This explains why the scores are lower than the  scores on the NIST 2009 test set.} The experiments are broken into
three
groups: baselines, syntax, and semantics.
To contextualize our results we experimented with a number of different
baselines that were composed from two different approaches to
statistical machine translation---phrase-based and hierarchical
phrase-based SMT---along with different combinations of language
model sizes and word aligners.  Our best performing baseline was a
Hiero model with a 5-gram language model and word alignments produced
using the Berkeley aligner.  The Bleu score for this baseline on the
development set was 23.1 Bleu points.

\begin{figure*}[t]
\begin{center}
\includegraphics[width=.95\linewidth]{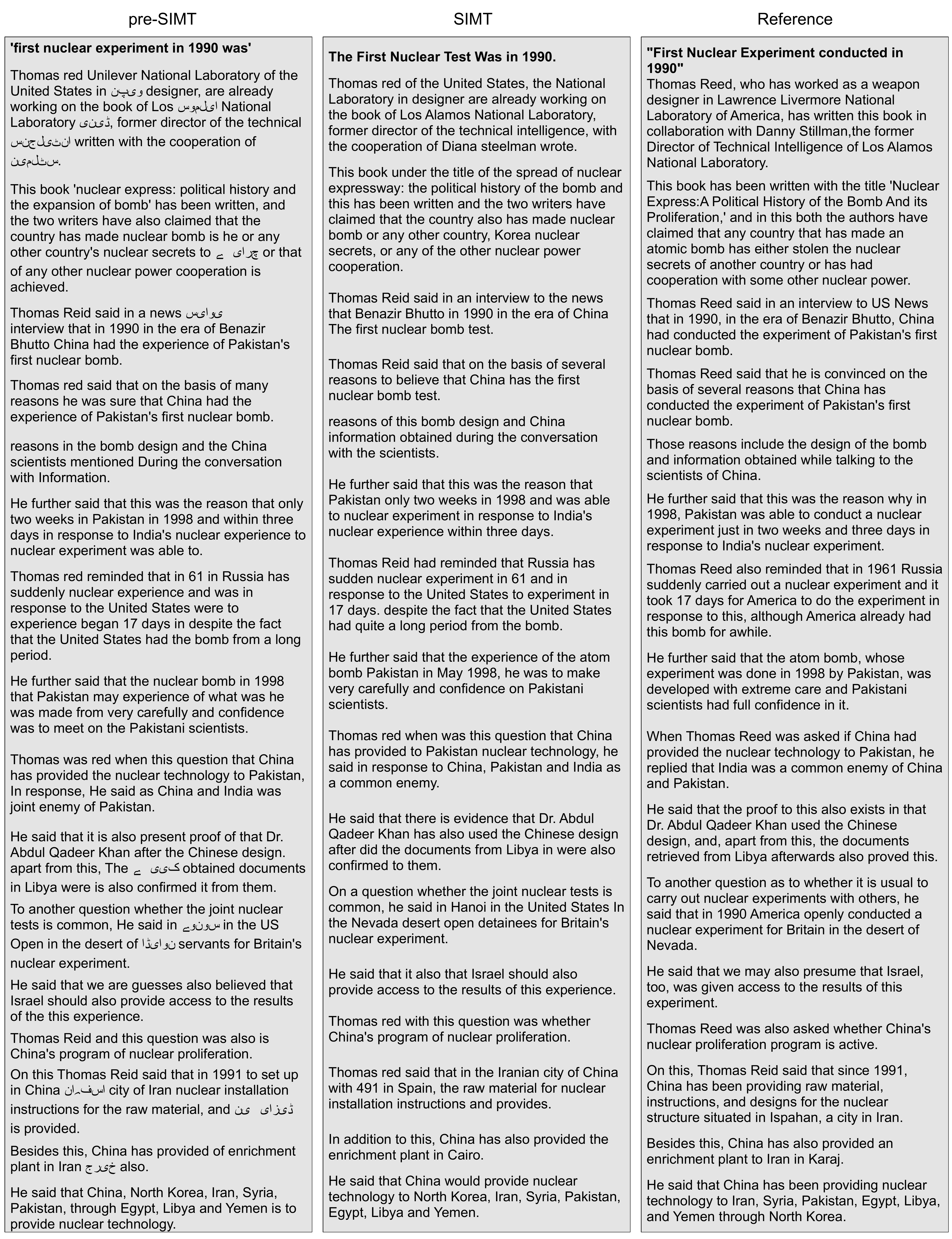}
\end{center}
\vspace*{-0.5cm}
\caption{An example of the improvements to Urdu-English translation before and after the SIMT effort. 
Output is from the baseline Hiero model, which does not use linguistic information, and from the final model, which incorporates 
syntactic and semantic information.}
\label{final-output}
\end{figure*}

After experimenting with syntactically motivated grammar rules, we
conducted three experiments on the effects of incorporating semantic
elements (e.g., named entities and modality markers) into the
translation grammars. 
In our devtest set our taggers tagged on average 3.5 named entities (NEs) per sentence and 0.35 modalities per sentence.
These were included by grafting NEs and modality markers onto the parse trees. 
Individually, each of
these made modest improvements over the syntactically-informed system
alone.  Grafting named entities onto the parse trees improved the Bleu
score by 0.2 points.  Modalities improved it by 0.3 points.  Doing
both simultaneously had an additive effect and resulted in a 0.5 Bleu
score improvement over syntax alone.  This improvement was the largest
improvement that we got from anything other than the move from
linguistically naive models to syntactically informed models.

Figure~\ref{final-output} shows example output from the final SIMT system.  
Notice that even in the title of the article, the SIMT system produces much more coherent English output than that of the 
linguistically naive system.
The figure also shows improvements due to transliteration, which are described in~\newcite{Irvine-PBML}. 
The scores reported in Figure~\ref{joshua-experiments} do not include transliteration improvements.

\section{Related Work}
\label{related-work}
This section describes related work in monolingual techniques for augmenting parsing, where parsing is applied to one language in the parallel text.

Our tree-grafting approach is related to a technique used for tree
augmentation in~\cite{Miller:2000}, where parse-tree nodes are
augmented with semantic categories. Miller et al.\@ augment tree nodes
with named entities and relations, while we used named entities and
modalities. The parser is subsequently retrained for both semantic and
syntactic processing.  The semantic annotations were done manually by
students following a set of guidelines and then merged with the
syntactic trees automatically.  In our work we tagged our corpus with
entities and modalities automatically and then grafted them onto the
syntactic trees automatically, for the purpose of training a
statistical machine translation system.  An added benefit of the
extracted translation rules is that they are capable of producing
semantically-tagged Urdu parses, despite that the training data were
processed by only an English parser and tagger.

Related work in syntax-based MT includes~\cite{HuangKnight06}, where
a series of syntax rules are applied to a source language string
to produce a target language phrase structure tree.  The Penn
English Treebank~\cite{Marcus:93} is used as the source for the syntactic
labels and syntax trees are relabeled to improve translation
quality.  In this work, node-internal and node-external
information is used to relabel nodes, similar to earlier work 
where structural context was used 
to relabel nodes in the parsing domain~\cite{KleinManning03}.
Klein and Manning's methods include lexicalizing determiners and percent markers,
making more fine-grained VP categories, and marking the properties of
sister nodes on nodes. All of these labels are derivable from the
trees themselves and not from an auxiliary source.

In the parsing domain, the work of~\cite{Petrov-Klein-2007:AAAI} is
related to the current work.  Petrov and Klein use a technique of rule
splitting and rule merging in order to refine parse trees during
machine learning.  Hierarchical splitting leads to the creation of
learned categories that have linguistic relevance, such as a breakdown
of a determiner category into two subcategories of determiners by
number, i.e., {\it{this}} and {\it{that}} group together as do
{\it{some}} and {\it{these}}.  We use rule splitting in cases where a
semantic category is inserted as a node in a parse tree, after the
English side of the corpus has been parsed by a statistical parser
(as described in section~\ref{algorithm}).

\section{Conclusions and Future Work}
\label{conclusions}

We have described a technique for translation that shows particular
promise for low-resource languages.  We have integrated linguistic
knowledge into statistical machine translation in a unified and
coherent framework.  We demonstrated that augmenting hierarchical
phrase-based translation rules with semantic labels (through
``grafting'') resulted in a 0.5 Bleu score improvement over syntax
alone.

Although our largest gains were from syntactic enrichments to the
Hiero model, demonstrating success on the integration of new semantic
aspects of language bodes well for future improvements based on the
incorporation of other
semantic aspects, e.g., relations and temporal knowledge, into
the translation rules, would further improve the translations.  The
syntactic framework is unique in its ability to support the
exploration of the impact of many different types of semantics on MT
quality.

Our findings indicate that the use of syntactic and semantic
information radically improves translation
quality for low-resource languages with different word order than
English.  Urdu has SOV (subject, object, verb) word order compared to
English SVO (subject, verb, object).  Thus, our observed improvements
are likely to be transferable to languages like Korean and Farsi, as
well as a host of other low-resource languages with different word
order.

The work presented here represents 
the first small steps toward a
full integration of MT and semantics.  Efforts underway in DARPA's
GALE program have already demonstrated the potential for combining MT
and semantics (termed {\it distillation\/}) to answer
the information needs of monolingual speakers using multilingual
sources. In previous work, however, semantic processing proceeded
largely independently of the MT system, operating only on the
translated output.  Our approach is significantly different in that it
combines syntax, semantics, and MT into a single model, offering the
potential advantages of joint modeling and joint decision-making.
It would be interesting to explore whether the integration of MT with
syntax and semantics can be extended to provide a single-model
solution for tasks such as cross-language information extraction and
question answering, and to evaluate our integrated approach, e.g., using
GALE distillation metrics.

\section*{Acknowledgments}

\vspace*{-.2in}
\renewcommand{\baselinestretch}{.95}
\small\normalsize
\hbox{\ }

We thank Aaron Phillips for help converting the output of the entity
tagger for ingest by the tree-grafting program.  We also thank Basis
Technology Corporation for their generous contribution of software
components to this work. This work is supported, in part, by the Johns
Hopkins Human Language Technology Center of Excellence, by
the National Science Foundation under grant IIS-0713448, and by
BBN Technologies under GALE DARPA/IPTO Contract
No. HR0011-06-C-0022. Any opinions, findings, and conclusions or
recommendations expressed in this material are those of the authors
and do not necessarily reflect the views of the sponsor.

\bibliographystyle{acl}
\bibliography{paper}

\end{document}